\documentclass[letterpaper]{article} 
\usepackage{aaai24}  
\usepackage{times}  
\usepackage{helvet}  
\usepackage{courier}  
\usepackage[hyphens]{url}  
\usepackage{graphicx} 
\usepackage{natbib}  
\usepackage{caption} 
\usepackage{algorithm}
\usepackage{algorithmic}

\usepackage{newfloat}
\usepackage{listings}
\DeclareCaptionStyle{ruled}{labelfont=normalfont,labelsep=colon,strut=off} 
\floatstyle{ruled}
\newfloat{listing}{tb}{lst}{}
\floatname{listing}{Listing}
\usepackage{lipsum}
\usepackage{amsmath}
\usepackage{amssymb}
\usepackage{enumitem}
\usepackage{booktabs}
\usepackage{multirow}
\usepackage{xcolor}
\title{Chain of Generation: \\ Multi-Modal Gesture Synthesis via Cascaded Conditional Control}
\author{
    Zunnan Xu \quad
    Yachao Zhang \footnotemark[2] \quad 
    Sicheng Yang \quad
    Ronghui Li \quad 
    Xiu Li\footnotemark[2]\\
}
\affiliations{
    Tsinghua Shenzhen International Graduate School, Tsinghua University\\
    University Town of Shenzhen, Nanshan District, 
    Shenzhen, Guangdong, P.R. China\\
    xzn23@mails.tsinghua.edu.cn, \{yachaozhang, li.xiu\}@sz.tsinghua.edu.cn
}

\usepackage{bibentry}

\begin{document}

\maketitle
\def\thefootnote{\dag}\footnotetext{Corresponding author}

\begin{abstract}
    This study aims to improve the generation of 3D gestures by utilizing multimodal information from human speech. Previous studies have focused on incorporating additional modalities to enhance the quality of generated gestures. However, these methods perform poorly when certain modalities are missing during inference. 
    To address this problem, we suggest using speech-derived multimodal priors to improve gesture generation. We introduce a novel method that separates priors from speech and employs multimodal priors as constraints for generating gestures. Our approach utilizes a chain-like modeling method to generate facial blendshapes, body movements, and hand gestures sequentially.
    Specifically, we incorporate rhythm cues derived from facial deformation and stylization prior based on speech emotions, into the process of generating gestures. By incorporating multimodal priors, our method improves the quality of generated gestures and eliminate the need for expensive setup preparation during inference.
    Extensive experiments and user studies confirm that our proposed approach achieves state-of-the-art performance.
\end{abstract}

\section{Introduction}
Gesture synthesis is a significant area of research within the realm of human-computer interaction (HCI), with diverse applications across various fields such as movies, robotics, virtual reality, and digital humans~\cite{kucherenko2021large}.
It is a challenging task that requires accounting for the dynamic movements of the human body, as well as the underlying rhythm, emotion, and intentionality~\cite{nyatsanga2023comprehensive}. 
In co-speech gesture generation, {three key indicators have emerged}: (i) generating gestures that synchronize with the audio and accurately depict the semantic content of the spoken text, (ii) generating gestures that are consistent with the speaker's style, and (iii) aligning the generated gestures with the speaker's intentions, including symbolic actions that may resemble sign language.

\begin{figure}
\begin{center}
\includegraphics[width=1.0\linewidth]{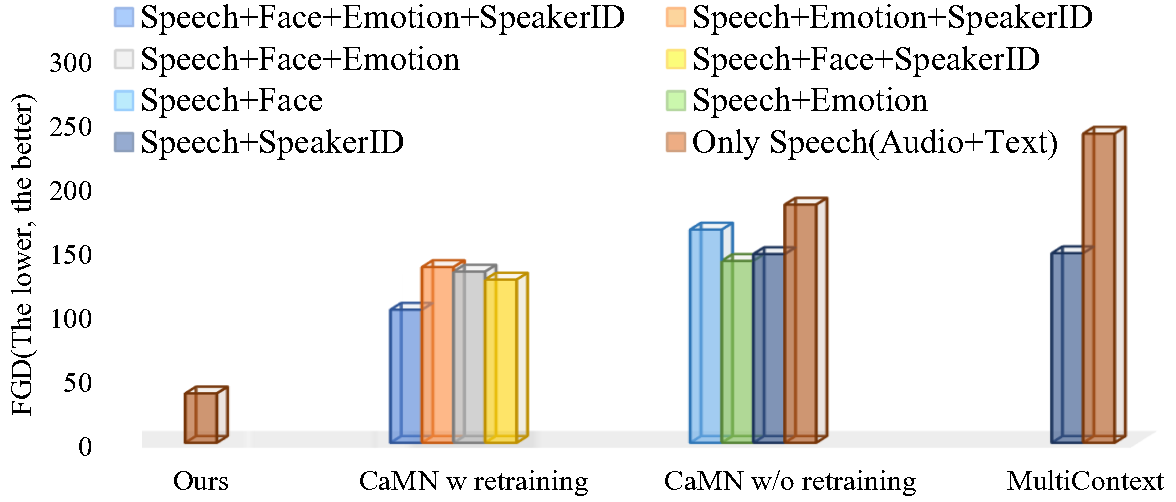}
\end{center}
   \caption{Performance comparison with limited modal during inference. The performance of existing multimodal methods is significantly hindered by the inadequate incorporation of multiple modalities during the inference stage. Our method addresses this limitation by utilizing prior information from speech to enable multimodal conditional control.
}  
\label{fig:intro}
\end{figure}

While substantial progress has been made in generating gestures synchronized to audio~\cite{ginosar2019learning,qian2021speech,yazdian2022gesture2vec,yang2023diffusestylegesture+,ao2023gesturediffuclip,yang2023unifiedgesture}, there has been limited exploration of emotive gesture generation that matches the rhythm and intention of the speech. 
Previous studies~\cite{yoon2020speech,liu2022beat} have demonstrated the effectiveness of introducing more modalities for gesture synthesis. 
However, most of these studies have not fully explored the potential of multimodal gesture synthesis modeling.
In CaMN~\cite{liu2022beat}, facial blendshapes and emotion labels are incorporated as additional inputs and represented as embeddings that {are} concatenated with the existing inputs to generate more native and expressive gestures.
However, as shown in Figure~\ref{fig:intro}, when the number of input modalities decreases, the performance of the model experiences a notable decline. 
In practical applications, it is common to encounter scenarios where some modalities are missing partially. Retraining a new model to accommodate these missing modalities can be a costly endeavor.

Recent studies~\cite{yang2023diffusestylegesture,qi2023emotiongesture} have explored incorporating emotion labels into the generation of human gestures using various techniques (e.g., random mask, cross attention), resulting in diverse and emotive gestures. 
However, these approaches still rely on using additional modalities, such as emotion labels, during the inference process. 
The inconsistency between the assigned emotions and the context of speech may also lead to unnatural gestures.
Furthermore, existing works overlook the importance of facial expressions in capturing speaker rhythm and intent, leading to suboptimal gesture generation.

\begin{figure*}[t]
\begin{center}
\includegraphics[width=1.0\linewidth]{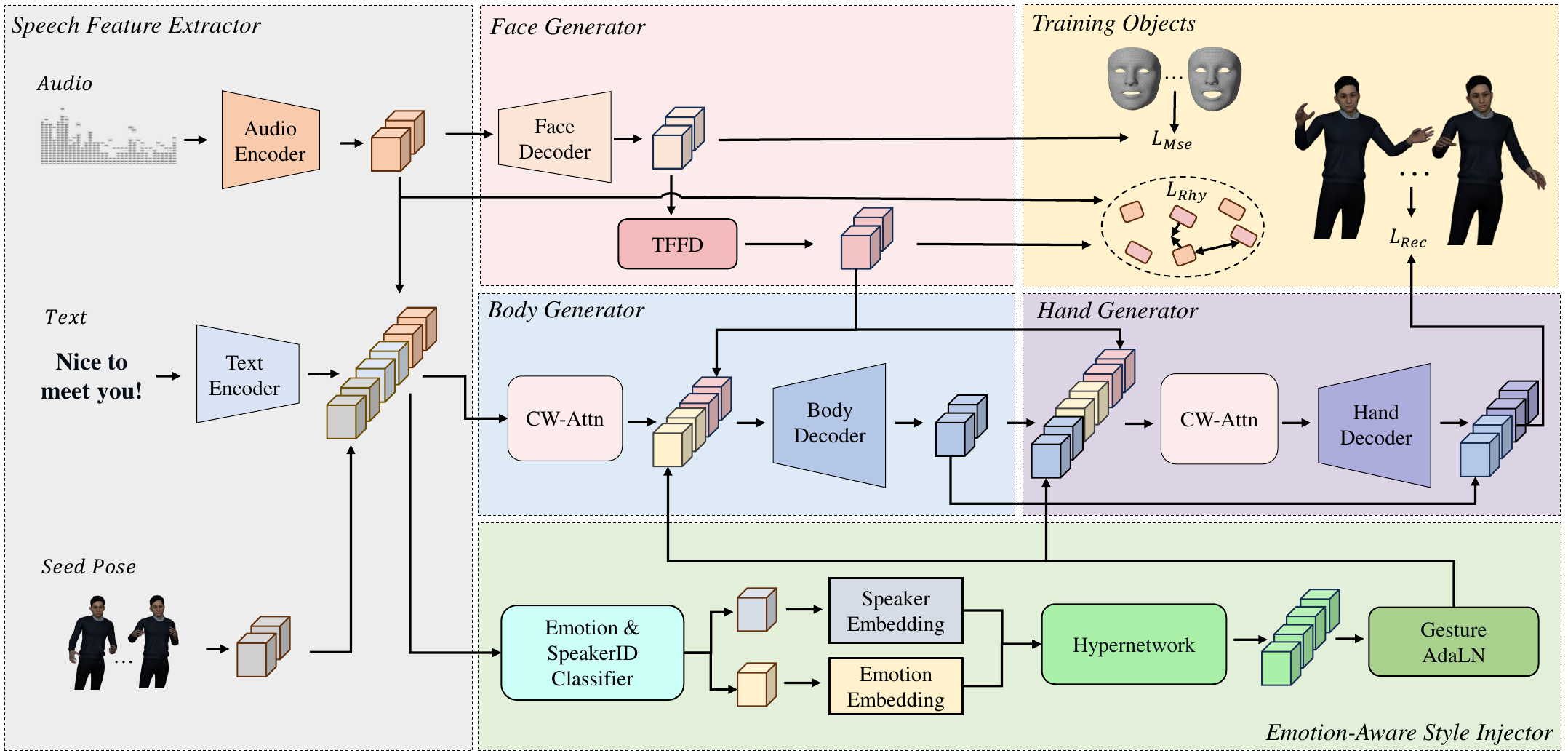}
\end{center}
   \caption{Overall architecture of the proposed CoG. Given audio sequences, text sequences, and seed poses, we use a chain-like framework to generate facial blendshapes, body movements, and hand gestures sequentially. We incorporate a classifier to extract emotion labels and speaker IDs from the speech, and use a hypernetwork to learn the style features. We further introduce gesture adaptive layer normalization to apply the generated style features to all input features used by the body and hand decoder. We leverage the ground truth of facial blendshapes as extra guidance during the training of the face generator to facilitate the transition from speech to facial blendshape. The features from the temporal facial feature decoder (TFFD) are utilized to calculate rhythmic identification losses, aiming to enhance the rhythm of generated gestures. 
   (Best viewed in color.) 
}  
\label{fig:arch}
\end{figure*}

 To address these problems, we propose a cascaded conditional control method for gesture synthesis. It aims to improve gesture generation by utilizing prior knowledge extracted from the speech, and maintain multimodal performance during inference when some modal data is missing (inference-efficient).
 Specifically, we introduce a novel method for extracting facial deformation and emotion-aware stylization priors from speech. By incorporating these priors, we greatly enhance the quality of gesture synthesis compared to previous methods. 
 We incorporate an emotion-aware style injector to highlight the importance of emotional expression within the larger speech context.
 Firstly, we train a classifier to extract emotion information from speech. The extracted emotion information is then transformed into style features. We further introduce gesture adaptive layer normalization to apply the generated style features to all input features used by the decoder, resulting in more expressive gestures.
 Meanwhile, we suggest using a face decoder to convert speech into facial blendshapes. Considering that facial blendshapes provide important prior knowledge of facial deformations (e.g., lip language, the rate of deformation related to speech rhythm), we propose a temporal facial feature decoder and a rhythmic identification loss to extract facial deformation-aware priors from speech, which helps generate more rhythmic gestures.
 To ensure consistency in the generated results, we propose a chain-like generation framework, as shown in Figure~\ref{fig:arch}, where the output of each stage serves as input for the next stage, ensuring a coherent generation across the face, body, and hands. 
 Our approach eliminates the need for costly preparation during inference. This is achieved by training the model to generate facial blendshapes and emotion information from speech during the training phase.
 Furthermore, by integrating these generated modalities into the cascaded generation process, our method improves the quality of the resulting emotional gestures. 
 The main contributions of our work are:
\begin{itemize}[leftmargin=*,noitemsep,nolistsep]
\item We propose a novel inference-efficient approach that enables the model to learn to extract multimodal prior information during training, eliminating the need for costly setup preparation (e.g., blendshape capture devices) during inference. 
Simplification of modalities can greatly reduce the difficulty of application, as in most cases, it is not guaranteed that all modalities can be collected.
\item We introduce a rhythmic identification loss to incorporate facial deformation as a guiding factor for generating gestures that synchronize with the speaker's speech rhythm.
\item To enhance emotional expression in generated gestures, we propose an emotion-aware style injector. This component extracts emotional information from speech and incorporates it as a stylization prior to gesture synthesis.
\item Extensive experiments and analyses demonstrate the effectiveness of the proposed approach.
\end{itemize}

\section{Related Work}
This work aims to develop a \textit{multimodal conditional} approach for \textit{co-speech gesture generation}.
In this section, we summarize previous studies and discuss the relations and differences.

\noindent\textbf{Multimodal Conditional Generation} aims to generate content conditioned on various input modalities. This requires models to understand the relationships between different modalities and use them to guide the generation process~\cite{qian2021speech}. Mainstream approaches for achieving multimodal fusion include:
(i) Attention mechanisms, which can model fine-grained inter-modality interactions~\cite{li2021audio2gestures,zhang2022motiondiffuse,li2023finedance};
(ii) Variational autoencoders, which can capture multimodal distributions~\cite{liu2022audio,qi2023emotiongesture};
(iii) Concatenation of representations from multiple modality-specific encoders~\cite{yoon2020speech,yang2022reprgesture}.
Recent studies~\cite{liu2022beat,yang2023diffusestylegesture} have included emotional embeddings as inputs. 
Some pioneering  works~\cite{qi2023emotiongesture,yin2023emog} achieve gesture diversity by learning emotion distributions and modifying emotion inputs during inference. 
However, these studies fail to consider the connection between emotion and the context in which speech occurs. Inconsistency between assigned emotions and the context of speech can result in unnatural gestures.
Moreover, these methods rely on a large amount of modalities as input. They suffer a significant drop in performance when the number of input modalities decreases.
Additionally, previous research has neglected the utilization of important priors that can be derived from facial blendshapes, leading to suboptimal gesture generation.
Our approach to multimodal conditional generation focuses on extracting multimodal priors from speech. 
This allows the model to efficiently utilize information from the speech context and generate emotional gestures that align with the speech rhythm.

\noindent\textbf{Co-speech Gesture Generation} focuses on generating gestures based on speech input. Previous methods can be categorized into three types:
(i) Linguistic rule-based methods convert speech into predefined gesture fragments and generate gestures using these rules~\cite{cassell1994animated, kopp2004synthesizing, wagner2014gesture};
(ii) Statistical models that learn mapping rules from data and combine them with predefined gesture units to generate gestures~\cite{kipp2007towards,levine2009real,levine2010gesture};
(iii) Deep learning methods that use neural networks to model the relationship between speech and gesture~\cite{yoon2019robots,yang2023QPGesture}.
While rule-based approaches can yield results that are easy to understand and control, they require a substantial amount of manual effort to create gesture datasets and engineer rules. Data-driven methods have become predominant for this task. Recent advances~\cite{hu2022uncertainty,psd} in deep learning have allowed neural networks  to directly learn the complex relationships between speech and gestures from raw multimodal data~\cite{zhang2022learning}. 
Previous research has shown that increasing the number of input modalities allows models to generate a wider range of expressive gestures~\cite{yoon2020speech,liu2022beat}.
Although incorporating additional input modalities (e.g., facial blendshapes, emotion) improved performance, it also led to greater application difficulty during inference (e.g., requiring more capture devices, longer preprocessing times).
Additionally, when there are multiple input modalities, there is often redundant modal information, allowing models to take shortcuts without fully utilizing all modalities.
To overcome these limitations, our method investigates the capacity of models to generate supplementary modalities exclusively from speech. This approach not only reduces inference costs but also enhances the utilization of multimodal features.

\section{Methodology}
\subsection{Overall Framework}
{To reduce the dependence on multimodal information that must be complete and consistent with training data during model prediction~\cite{lin2023consistent123}, we aim to design a gesture synthesis method that can utilize multimodal prior information for better training, while still maintaining high inference performance when lacks partial modal information. Such a modal that eliminates the costly setup preparation during inference, can improve its flexibility and applicability. We propose a chain of generation method: Cascaded Gesture Synthesizer, which models the gesture in order from simple to complex, beginning with the facial blendshape, followed by the body, and concluding with the hand.} 

{Specifically, as a common approach in speech-driven gesture synthesis, we use audio sequences $A=\{a_1,...,a_N\}$, text sequences $T=\{t_1,...,t_N\}$ and previous gestures as input (e.g. seed poses) to guide the continuous sequence of co-speech gestures denoted as $G=\{g_1,...,g_N\}$,  as inputs. Here, $N$ represents the total number of frames, and $g_i\in R^{J\times 3}$ represents the 3D rotation pose state of the $i$-th frame. }
For speech input words, we use a pre-trained FastText~\cite{bojanowski2017enriching} to convert them into a word embedding set. Then, the word sets are fine-tuned by an encoder $E_T$ to generate the text feature $T\in \mathbb{R}^{128}$. For audio feature extraction, we utilize a semi-supervised speech model, wav2vec2.0~\cite{baevski2020wav2vec}. 

{We found that it is difficult to maintain the performance during the inference phase if some modal information is missing. 
Therefore, we first introduce a linear projection layer at the end of the audio encoder to obtain the 128-dimensional latent audio feature $A \in \mathbb{R}^{128}$, for refining the audio feature. In order to guide the transition, we present a supervision signal (annotated facial blendshapes) during training to encourage the separation of facial blendshapes from speech. Then, we also extract rhythmic features from facial deformation using a temporal convolutional decoder and a rhythmic identification loss. Finally, to ensure accurate emotional expressions in the generated poses, we propose an emotion-aware style injector, which incorporates an additional classifier that generates emotion information from the speech input, and a hypernetwork to learn the style features. We further introduce gesture adaptive layer normalization to apply the generated style features to all input features used by the body and hand decoder. }
In this way, we {utilize} the deformation prior of facial blendshapes and the stylization prior derived from speech to improve {gesture} synthesis.

\subsection{Cascaded Gesture Synthesizer} \label{sec3:Syn}
Inspired by the concept of chain of thought and modal interaction~\cite{wei2022chain, xu2023bridging}, we propose to decompose the  problem of speech-to-gesture generation into several stages. These stages include speech-to-emotion-based style recognition, speech-to-facial blendshape generation, facial blendshape-to-body generation, and body-to-hands generation.
With emotion-based style features integrated into the latent features, our approach utilizes a chain-like modeling method, sequentially generating facial blendshapes, body movements, and hand gestures. 

\noindent\textbf{Face Generator.}
Our face generator is designed to generate a sequence of facial blendshapes $F=\{f_1,...,f_N\}$ that are synchronized with audio features $A=\{a_1,...,a_N\}$, where $N$ denotes the total number of frames. To achieve this goal, we formulate 3D facial blendshape synthesis as a speech-driven sequence-to-sequence learning problem. 
Specifically, we adopt a 3-layer temporal convolutional decoder {( $\text{FaceDec}(\cdot)$) to transfer audio features into facial blendshape $F_t$. 
Considering the temporal transformation of facial blendshape (e.g., the rate and degree of deformation), which can help the model understand the rhythm of speech, we decode the blendshapes into facial features $\hat{F}_t$ using a temporal facial feature decoder ($\text{TFFDec}(\cdot)$}), which consists of 4 layers of temporal convolutional layers. The features will be further used to generate gestures in the body and hand generator. This allows us to capture local facial deformation over time, which can be formalized as: 
\begin{equation}
\begin{aligned}
& F_t=\text{FaceDec}({A}_t), \\
& \hat{F}_t=\text{TFFDec}({F}_{t}),
\end{aligned}
\label{FBG}
\end{equation}
where $t\in N$ denotes the {t}-th frame and $N$ represents the total number of frames.
 Due to the temporal alignment of the annotated blendshape and speech, we add supervision on $F_t$ to ensure the time consistency of the generated facial features with the speech. 

{
\noindent\textbf{Body and Hand Generator.}
As two generators have the same structure, we use body generator as an example to introduce this method for simplicity. In this generator, two modules (Channel-wise Attention Module and Body/Hand Decoer) are introduced detailed as follows. }

\textbf{Channel-wise Attention Module (CW-Attn).} 
In order to facilitate the learning of features, we introduce the channel-wise attention module to perform adaptive reweighting of channel information to focus on more salient features. These attention weights selectively enhance informative channels via channel-wise multiplication of the latent features, which we refer to as ``attended features''.
This module emphasizes useful channels while suppressing less informative ones. It contains average and max pooling layers to aggregate channel-wise statistics. The pooled outputs are fed into convolutional layers to generate a per-channel attention vector, activated by a sigmoid to range from 0 to 1, formulated as{ $\text{CW-Attn}(\cdot)$.
We concatenate the multi-modal features and feed them into CW-Attn as: 
\begin{equation}
C_t =\text{CW-Attn}([\hat{A}_t:\hat{T}_t:\hat{F}_t]),
\end{equation}
where $t\in N$ denotes the i-th frame, $N$ represents the total number of frames, and $[:]$ denotes the concatenation operation.  The attention vector is generated by convolving the pooled features, which selectively enhances salient channels via element-wise multiplication with the input.}

\textbf{Body Decoder.} We adopt the separated, cascaded LSTM structure from previous work~\cite{liu2022beat} {as decoder to capture the body/hand feature. Unlike these} approaches, we incorporate multimodal priors as supplementary conditions to enhance gesture naturalness. 

For reconstruction, the independent MLPs are added to the end of the body decoders to enable body synthesis. The process can be formalized as:
\begin{equation}
\begin{aligned}
M_t &=\text{StyleInject}(C_t,S_t), \\
\hat{B}_t &=\text{BodyDec}(M_t) ,\\
B_t &=\text{BodyMlp}(\hat{B}_t) .
\end{aligned}
\label{BGG1}
\end{equation}
where $\text{StyleInject}$ refers to the operation of injecting emotive and personal gesture style, which is given in (\S\ref{sec3:style}). 

\textbf{Hand Decoder.} {The difference is that the input feature of the hand generator concatenates the output of the body generator, denoted as:
\begin{equation}
    \hat{C}_t =\text{CW-Attn}([M_t:\hat{B}_t]).
\end{equation} 
We further model the hand gesture synthesis and get the $\hat{M}_t$, $\hat{H}_t$, and $H_t$ same as body decoder.}

\subsection{Emotion-Aware Style Injector} \label{sec3:style}
We introduce an emotion-aware style injector to enable more expressive gestures. 
Contrary to previous works~\cite{liu2022beat,yang2023diffusestylegesture} that simply converts emotional labels into embedded features and concatenates them with other modal features as inputs. We consider emotional gesture generation as a stylized task and utilize a classifier to derive emotional information from speech.

\noindent\textbf{Emotion \& Speaker ID Classifier.} 
To extract emotional information from the speech input, we first train a classifier on the training set using speech, emotion labels and speaker IDs. The classifier consists of a 3-layer temporal convolutional network with two linear projection layers, enabling the prediction of probabilities.
We utilize cross-entropy loss to optimize the alignment between the predicted probability and the true emotion and speaker category. Then, we freeze the weights of the classifier and use its predictions as a guide to learn the style vector for the cascaded gesture synthesizer.

\noindent\textbf{Gesture Adaptive Layer Normalization.}
We build upon the concept of adaptive layer normalization~\cite{huang2017arbitrary} and propose GestureAdaLN for stylized gesture generation. 
GestureAdaLN utilizes a hypernetwork to leverage inter-speaker and inter-emotion priors. The hypernetwork takes speaker and emotion embeddings as inputs and employs a 2-layer temporal convolution network to generate a style vector $S_t$ for representing gesture style. Through two linear projection layers ($f(\cdot)$ and $g(\cdot)$), the style vector $S_t$ is mapped to channel-wise mean and standard deviation parameters. These parameters are then used to modulate the latent feature $X$.
The process can be formalized as:
\begin{equation}
\begin{aligned}
S_t &= \text{HyNet}(E_t, I_t),  \\
\hat{X} &= f(S_t) \cdot X + g(S_t),
\end{aligned}
\label{GesAdaLN}
\end{equation}
where $\text{HyNet}(\cdot)$ denotes the hypernetwork, and $E_t$ and $I_t$ represent the emotion labels and speaker ids at frame $t$. 
By considering the speaker identity and emotions, our method effectively captures the style differences in gesture content across different contexts, resulting in gestures that better align with the current speech content.

\subsection{Training Objective} \label{sec3:objective}
\noindent\textbf{Rhythmic Identification Loss.} 
Given the significance of a speaker's speech rhythm in their nonverbal communication, we suggest a novel method to aid in the generation of gestures from audio by integrating speech rhythms. A key insight is that the rhythm and prosody of speech provide important cues for generating natural gestures, while facial blendshapes provide additional contextual information about the tone and intent.
Therefore, we apply the InfoNCE loss~\cite{chen2020simple} to encourage temporal synchronization between the facial features $\hat{F}$ and audio features $\hat{A}$ to extract the speaking rhythm. Specifically, we compute the InfoNCE loss between encoded representations of the two modalities, which are obtained using neural network encoders $f$ and $g$, respectively. This acts as a multimodal alignment loss that matches rhythmic cues in the audio with facial expressions, thereby generating better features with rhythmic information.
The loss is defined as:
\begin{equation}
\ell_{\text{Rhy}} = -\frac{1}{N}\sum_{i=1}^{N}\log \frac{\exp\left(\mathrm{sim}(f(\hat{F}_i), g(\hat{A}i))/\tau\right)}{{\sum_{j=1}^{N}}\exp\left(\mathrm{sim}(f(\hat{F}_i), g(\hat{A}_j))/\tau\right)},
\end{equation}
where $N$ is the number of frames in the aligned sequences, and $\hat{F}_i$ and $\hat{A}_i$ are the facial and audio features at the i-th frame, respectively. $\mathrm{sim}(\cdot)$ denotes cosine similarity between the encoded representations, and $\tau$ represents the temperature hyperparameter. 
The synchronization loss encourages neural network encoders to learn representations that capture meaningful correlations between two modalities while disregarding irrelevant variations from the speech.

\noindent\textbf{Face Reconstruction Loss.} 
We utilize the mean squared error (MSE) loss as the face blendshape reconstruction loss. Specifically, the loss is defined as:
\begin{equation}
\ell_{\text{MSE}} = \frac{1}{N}\sum_{i=1}^{N}(\mathbf{F}_i - \hat{\mathbf{F}}_i)^2,
\end{equation}
where $N$ indicates the number of frames, $\mathbf{F}_i$ denotes the $i$-th frame of ground truth blendshape parameters, and $\hat{\mathbf{F}}_i$ denotes the corresponding predicted blendshapes by our model. By minimizing the MSE during training, we aim to improve the accuracy and fidelity of the facial blendshapes in capturing facial expressions and deformations. This, in turn, enables the generation of expressive gestures by cascaded gesture synthesizer. 

\noindent\textbf{Body \& Gesture Reconstruction Loss.} 
For body and gesture generator, we utilize the L1 loss as the reconstruction loss function. The L1 loss measures the absolute difference between the predicted and ground truth values of the body and gesture parameters, providing a robust and efficient metric for reconstruction quality. 
The losses are defined as:
\begin{equation}
\begin{aligned}
\ell_{\text{rec}}^B = \mathbb{E}\left[\left\|\mathbf{B} - \hat{\mathbf{B}}\right\|_1\right], & ~
\ell_{\text{rec}}^H = \mathbb{E}\left[\left\|\mathbf{H} - \hat{\mathbf{H}}\right\|_1\right], \\
\ell_{\text{Rec}} = &\ell_{\text{rec}}^B + \alpha \ell_{\text{rec}}^H,
\end{aligned}
\end{equation}
where a weight $\alpha$ is adopted to balance the body and hands penalties, and $\mathbb{E}$ indicates the maximum likelihood estimation.
Our objective is to improve the quality of the generated gesture by minimizing the L1 loss during training. This, in turn, enhances the model's capability to capture the dynamics and subtle nuances of human motion.

\noindent\textbf{The Overall Objective.} In summary, the overall optimization objective for our proposed method is formalized as:
\begin{equation}
\mathcal{L}=\lambda_{\text{Rhy}} \ell_{\text{Rhy}}+\lambda_{\text{Mse}}\ell_{\text{Mse}}+\lambda_{\text{Rec}}\ell_{\text{Rec}}.
\end{equation}
where $\lambda_{\text{Rhy}}=1$, $\lambda_{\text{MSE}}=1000$ and $\lambda_{\text{Rec}}=500$.

\section{Experiments}
\subsection{Experiments Setting}
\noindent\textbf{Dataset.}
To evaluate the effectiveness of each component in our approach, we conducted comprehensive experiments on a large-scale multimodal dataset called BEAT (Body-Expression-Audio-Text)~\cite{liu2022beat}. The dataset comprises 76 hours of multi-modal data captured from 30 speakers conversing in four different languages while expressing eight distinct emotions. This includes conversational gestures accompanied by facial expressions, emotions, and semantics, as well as annotations for audio, text, and speaker identity.
To ensure a fair comparison, we followed CaMN~\cite{liu2022beat} and utilized approximately 16 hours of speech data from English speakers. Additionally, we followed the established practice of dividing the dataset into separate training, validation, and testing subsets, while maintaining the same data partitioning scheme as in previous work to ensure the fairness of the comparison.

\noindent\textbf{Implementation Details.}
We use the Adam optimizer with an initial learning rate of 0.00025, and set the batch size to 512. To ensure a fair comparison, we use N = 34 frame clips with a stride of 10 during training. The initial four frames are used as seed poses, and the model is trained to generate the remaining 30 poses, which correspond to a duration of 2 seconds. Our models utilize 47 joints in the BEAT dataset, including 38 hand joints and 9 body joints. The latent dimensions of the facial blendshape, audio, text, and gesture features are all set to 128, while the speaker embedding and emotion embedding are set to 8. We set $\tau=0.1$ in the rhythmic identification loss.
All experiments are conducted using NVIDIA A100 GPUs. 
\textit{More analysis results about hyperparameter are given in supplementary materials. }

\subsection{Evaluation Metrics.}
\noindent\textbf{Fréchet Gesture Distance (FGD)}
We used the Fréchet Gesture Distance~\cite{yoon2020speech} to evaluate the distribution distance between the synthesized and ground truth gestures. To compute this metric, we utilize the autoencoder pre-trained by BEAT~\cite{liu2022beat}.

\noindent\textbf{Semantic-Relevant Gesture Recall (SRGR).}
We adopt the Semantic-Relevant Gesture Recall metric~\cite{liu2022beat} to evaluate the semantic relevance of generated gestures. SRGR leverages the semantic scores as weights for the Probability of Correct Keypoint metric between the generated gestures and the ground truth gestures. It can accurately capture the semantic aspects related to the generated gestures.

\noindent\textbf{Beat Alignment Score (BeatAlign).}
To evaluate the correlation between gestures and audio, we utilized the Beat Alignment Score~\cite{li2021ai} to calculate the similarity between gesture beats and audio beats. BeatAlign provides a measure of alignment between the two modalities.

\subsection{Qualitative Results}
We compared our proposed method with existing multi-modal gesture synthesis methods on the BEAT dataset in Table \ref{table:rsota}.
Our approach achieves competitive performance across all metrics when compared to the state-of-the-art methods, which validates the effectiveness of our cascaded conditional framework for this task.
In our method, we adopt an incremental modeling approach for human gestures. We start by generating facial blendshapes, then move on to the body, and finally the gestures. 
The chain-like generation pipeline enables lower FGD, leading to high-fidelity gesture reconstruction by gradually modeling gestures from simple to complex. 
Additionally, we effectively utilize multimodal information by decoupling stylization and rhythm cues from speech, resulting in significant improvements in the SRGR and BeatAlign scores.

\begin{table}
\caption{Comparison with state-of-the-art method in the term of FGD, SRGR and BeatAlign. All methods are trained on BEAT datasets. 
$\downarrow$ denotes the lower the better while $\uparrow$ denotes the higher the better. The best results are in bold. 
}
\centering
\footnotesize
\label{table:rsota}
\resizebox{\columnwidth}{!}{%
\begin{tabular}{l|ccc}
    \toprule
    Methods & $\text{FGD}$ $\downarrow$ & $\text{SRGR}$ $\uparrow$  & $\text{BeatAlign}$ $\uparrow$ \\
    \midrule
    Seq2Seq~\cite{yoon2019robots} & 261.3 & 0.173 & 0.729 \\ 
    Speech2Gesture~\cite{ginosar2019learning} & 256.7 & 0.092 & 0.751  \\
    MultiContext~\cite{yoon2020speech} & 176.2 & 0.195 & 0.776 \\
    Audio2Gesture~\cite{li2021audio2gestures} & 223.8 & 0.097 & 0.766  \\
    CaMN~\cite{liu2022beat} & 123.7 & 0.239 & 0.783  \\
    TalkShow~\cite{yi2022generating}  & 91.00 & - & 0.840 \\
    GestureDiffuCLIP~\cite{ao2023gesturediffuclip}  & 85.17 & - & - \\ 
    \textbf{CoG (ours)}  & \textbf{45.87} & \textbf{0.308}  & \textbf{0.931}  \\
    \bottomrule
\end{tabular}
}
\end{table}

\begin{figure}
\begin{center}
\includegraphics[width=1.0\linewidth]{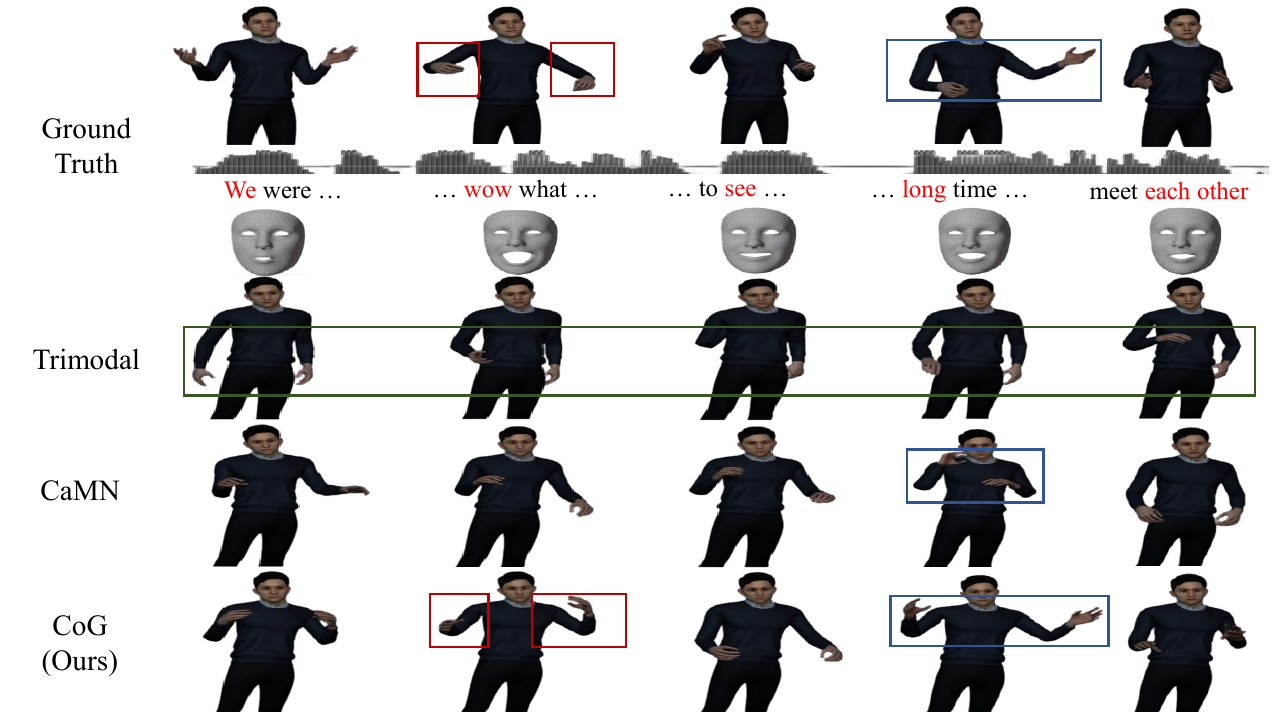}
\end{center}
   \caption{Visualization of our predicted 3D gestures against various baseline methods. The generated results of different methods are presented in separate rows, with each row representing the generated results of a method at different time frames. 
} 
\label{fig:vis}
\end{figure}

\subsection{Ablation Study}
We validate the effectiveness of our proposed approach by conducting ablation studies on different components of our proposed method.

\noindent\textbf{Effect of Cascaded Gesture Synthesizer.} 
We evaluated the effectiveness of the cascaded gesture synthesizer through experiments conducted under various settings, as shown in Table~\ref{table:ablation_cog}.
We examined the significance of the chain structure by conducting ablative experiments on the face generator.  
Specifically, ``$-\,\text{face}_{cog}$'' refers to the method that does not join the face generator in our cascaded framework. In this setting, the facial blendshapes are converted into embeddings using a facial encoder and then concatenated with the encoding features of other modalities.
``$+\,\text{face}_{cog}$'' represents the method that incorporates a face generator, where facial blendshape is included as a generation condition using the temporal facial feature decoder. 
The results of the ablation study show that incorporating the temporal facial feature decoder to process the generated features leads to a notable 31.7\% decrease in FGD.
We also examine the impact of integrating a channel-wise attention module, which highlights important channels while suppressing less informative ones. By incorporating the channel-wise attention module, indicated as "+ cw-attn," we achieve adaptive reweighting of channel information, giving higher priority to more prominent features and leading to enhanced overall metrics.

\begin{table}[b]
\centering
\caption{Ablation study on different components of our proposed method. $\downarrow$ denotes the lower the better, and $\uparrow$ denotes the higher the better.}
\label{table:ablation_cog}
\footnotesize
\begin{tabular}{lcccccc}
\toprule
\begin{tabular}[c]{@{}c@{}}Settings\end{tabular} & FGD$\downarrow$ & SRGR$\uparrow$ & BeatAlign$\uparrow$ \\ 
\midrule
$-\,\text{face}_{cog}$  & 88.50 & 0.228 & 0.762   \\
$+\,\text{face}_{cog}$  & 60.44 & 0.240 & 0.834  \\
$+\,\text{cw-attn}$  & 58.74 & 0.238 & 0.842   \\
$+\,\text{style}$  & 52.36 & 0.241 & 0.916 \\
$+\,\mathcal{L}_{Rhy}$  & 45.87 & 0.308 & 0.931 \\
\bottomrule
\end{tabular}%
\end{table}

\noindent\textbf{Effect of Emotion-Aware Style Injector.}
We validated the effectiveness of the emotion-aware style injector, as shown in Table~\ref{table:ablation_cog}, the method labeled as ``$+\,\text{style}$'' includes the emotion-aware style injector to enable more expressive gestures. 
The ablation results demonstrate that incorporating emotional information from speech as a stylized prior enhances our method's capability to generate natural and contextually appropriate gestures. 
Moreover, the results confirm that considering gesture synthesis as a stylized task can enhance the expressiveness of the generated gestures. This improvement is clearly demonstrated in the overall metrics, particularly in the case of BeatAlign, which experienced a significant increase of 8.79\%.

\noindent\textbf{Effect of Rhythmic Identification Loss.} 
We validate the effectiveness of the rhythmic identification loss, as listed in Table~\ref{table:ablation_cog}, ``$+\,\mathcal{L}_{Rhy}$'' signifies the method that incorporates the rhythmic identification loss for further exploration of the rhythm of gesture. 
This improvement provides evidence for the correlation between facial blendshapes and gesture rhythm, as well as the successful separation of speech and facial features through contrastive learning. The benefits of investigating rhythm through facial features are clearly illustrated by the significant 27.8\% increase in the SRGR score.

\subsection{Qualitative Analysis}
\noindent\textbf{User Study.}
We conducted user study to evaluate the visual quality of the generated co-speech 3D gestures. 
For each compared method, we generated 10 results, and these gestures were converted into videos for evaluation by 22 participants. 
In each test, participants are presented with 20-second video clips synthesized by different models.
Then, the participants are asked to provide ratings based on four dimensions: (i) naturalness, (ii) appropriateness, (iii) style correctness, and (iv) synchrony.
In the naturalness test, participants are asked to evaluate the similarity of the generated gestures to those made by humans. This assessment primarily focuses on the naturalness and smoothness of the movements.
In the appropriateness test, participants are required to assess the consistency of the gestures with the speech content, including both the literal content and the conveyed semantic meaning.
In the style correctness test, participants are provided with emotion labels and asked to determine whether the generated gestures align with the intended style.
In the synchrony test, participants assess the level of synchronization between gestures, speech rhythm, and accompanying audio and facial movements. They evaluate how well the gestures synchronize with these elements, ensuring a cohesive and harmonious overall presentation.
We compared three methods, including MultiContext~\cite{yoon2020speech}, CaMN~\cite{liu2022beat}, our method, and ground truth.
As shown in Table~\ref{table:user}, the average results demonstrate that our method achieves a significant advantage over the compared methods, performing better across all metrics.

\begin{table}[b]
\centering
\caption{The user study on \textbf{naturalness} (human likeness), \textbf{appropriateness}  (the degree of consistency with the speech content), style \textbf{correctness} (with emotion labels), and \textbf{synchrony} (the level of synchronization with the speech rhythm). The rating score range is 1-5, with 5 being the best. $\uparrow$ indicates the higher the better.}
\label{table:user}
\resizebox{\columnwidth}{!}{%
\begin{tabular}{lcccccccccc}
\toprule
Methods & Naturalness$\uparrow$  & Matchness$\uparrow$ & Correctness$\uparrow$ & Synchrony$\uparrow$ \\ 
\midrule
MultiContext & 3.127 & 2.901 & 3.129 & 3.110   \\ 
CaMN & 3.588 & 3.261 & 3.225 & 3.414  \\ 
\textbf{Ours} & 3.916 & 3.624 & 3.541 & 3.685   \\
Ground Truth & 4.492 & 4.570 & 4.382 & 4.413   \\
\bottomrule
\end{tabular}%
}
\end{table}

\begin{figure}[t]
\begin{center}
\includegraphics[width=1.0\linewidth]{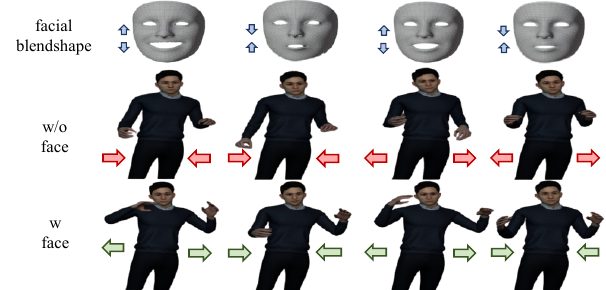}
\end{center}
   \caption{Visualization of the gestures generated by our method, the results of without/with a facial generator and rhythmic identification loss (denoted as \textit{w/o face} and \textit{w face}). The facial blendshape results are shown in the first row, and \textit{w/o face} and \textit{w face} are given in the second and third rows, respectively. 
} 
\label{fig:vis2}
\end{figure}

\noindent\textbf{Visualization.}
As illustrated in Figure~\ref{fig:vis}, our method generates gestures that are more rhythmic and natural, aligning well with the speaking cadence. 
The gestures highlighted within the green rectangle indicate that previous methods were lacking in terms of diversity. In contrast, the gestures generated by our method exhibit a greater range of diversity.
The gestures circled within the red rectangle indicate that our method is capable of recognizing gestures corresponding to expressive words. As shown in the figure, when the speaker utters an expressive word like ``wow'', the corresponding audio exhibits rhythmic fluctuations. Our method responds by performing a simultaneous upward movement of both hands, resembling the human gestures represented by the ground truth. In contrast, previous methods have been deficient in capturing this aspect.
The gestures highlighted within the blue rectangle reveal that our method is capable of producing gestures that correspond to the intended meaning of the speaker. It can be observed that when the speaker mentions the word ``long'', our method generates a gesture of both hands spreading outwards, which aligns with the semantic meaning of the word. This gesture closely resembles the ground truth. In contrast, other methods lack a similar response in this regard.    
Our approach enables the generation of expressive gestures that not only appear natural but also synchronize with the rhythm of speech.

We compared the results generated without the face generator to those generated with the face generator in Figure~\ref{fig:vis2}. It can be observed that incorporating facial features aligns gesture rhythms with speech pace, demonstrating the advantages gained from including facial deformation priors.

\section{Conclusion}
In this study, we propose a framework to enhance the generation of 3D gestures by leveraging multimodal information from human speech. Our approach incorporates multimodal priors as constraints to enhance gesture generation. 
We adopt a chain-like modeling approach to sequentially generate facial blendshapes, body movements, and hand gestures.
By incorporating rhythm cues from facial blendshapes and stylization priors into the generation process, our approach improves the  quality of the generated gestures and reduces the number of modalities needed during inference.

\section*{Acknowledgements}
This research was partly supported by Shenzhen Key Laboratory of next generation interactive media innovative technology (Grant No: ZDSYS20210623092001004), the China Postdoctoral Science Foundation (No.2023M731957), the National Natural Science Foundation of China under Grant 62306165.

\bibliography{aaai24}
\newpage
\appendix
\section{Implementary Details}
\subsection{BEAT Datasets}
\begin{figure}[b]
\begin{center}
\includegraphics[width=1.0\linewidth]{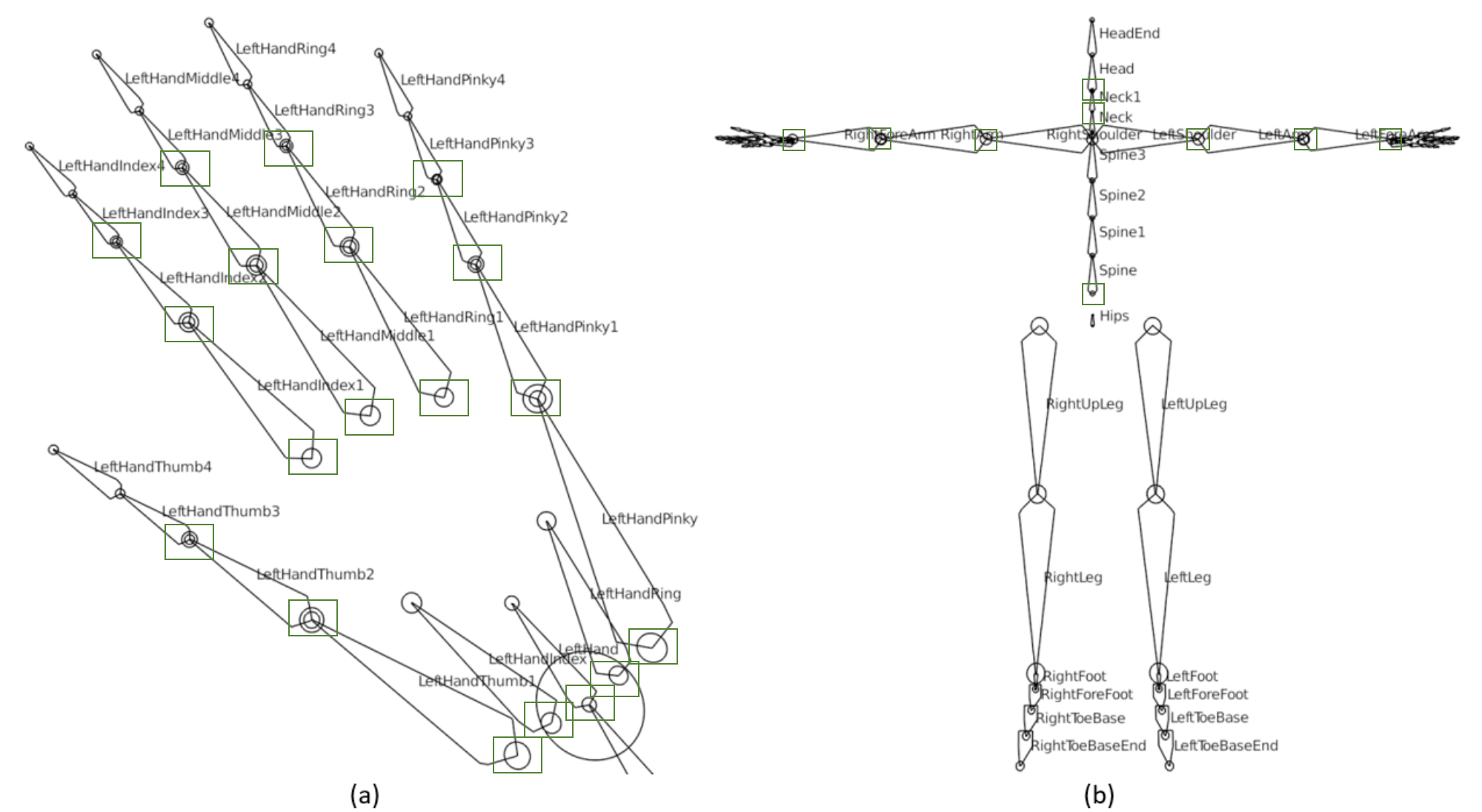}
\end{center}
   \caption{The joints of the upper body and hands utilized in our method. Our models utilize 47 joints in BEAT dataset, including 38 hand joints and 9 body joints.
} 
\label{fig:joint}
\end{figure}

Given that our model requires annotated facial blendshapes for speech-to-blendshape conversion during training, as well as labels for emotion and speaker ID for classifier training, we have selected the BEAT  (Body-Expression-Audio-Text)~\cite{liu2022beat} datasets as our experimental dataset.
The dataset consists of 76 hours of multi-modal data collected from 30 speakers expressing eight different emotions.
This includes nonverbal communication through gestures, facial expressions, emotions, and semantics, as well as annotations for audio, text, and speaker identification.
Specifically, the types of emotions include neutral, happiness, anger, sadness, contempt, surprise, fear, and disgust. We follow previous research and use around 16 hours of speech data from four English speakers. 
The BEAT dataset includes a total of 48 hand joints and 27 body joints in the human skeleton. 
Following previous work, we only use select joints shown in Figure~\ref{fig:joint}, as our synthesis focuses on the upper body.

\subsection{Audio Encoder}
In order to make audio feature extraction, we utilize a semi-supervised speech model, wav2vec2.0~\cite{baevski2020wav2vec}, which was pretrained and fine-tuned on 960 hours of Librispeech on 16kHz sampled speech audio. 
The encoder is composed of an audio feature extractor and a transformer encoder, resulting in a 768-dimensional representation of the speech. 
We freeze the parameters of the feature extractor. 
For audio feature refinement and alignment, a linear projection layer is applied at the end of the encoder to obtain the final 128-dimensional latent audio feature $A \in \mathbb{R}^{128}$.

\subsection{Text Encoder}
For speech input words, we utilize a pre-trained FastText~\cite{bojanowski2017enriching} to convert them into a word embedding set $V^T\in \mathbb{R}^{300}$. Then, the word sets are fine-tuned by a customized encoder $E_T$, which is an 8-layer temporal convolution network (TCN) with skip connections.
Since the model was trained to generate 30 poses over a 2-second period (15 fps) using the first 4 frames as seed poses, the TCN fuses the information from 2f = 34 frames to generate the final latent feature of the text. The set of features is denoted as $T\in \mathbb{R}^{128}$.

\subsection{Emotion \& Speaker ID Classifier}
We choose to use a classifier to guide gesture style based on the assumption that people with similar speaking styles (e.g., tone of voice, emotional expressiveness) will also have similar gesture styles.
The classifier consists of a 3-layer temporal convolutional network, with two linear projection layers at the end of the backbone to enable probability prediction.
Before training the CoG framework, we initially train the classifier on the training set using speech, emotion labels, and speaker IDs.
We use cross entropy loss to optimize the alignment between the predicted probability and the true emotion and speaker category.
Specifically, there are eight emotions and four speakers, and the losses are defined as:
\begin{equation}
\begin{aligned}
\ell_{\text{emo}} &= H(P_{emo}, Q_{emo}), \\
\ell_{\text{id}}~ &= H(P_{id}, Q_{id}), \\
\ell_{\text{cls}} &= \ell_{\text{emo}} + \ell_{\text{id}}.
\end{aligned}
\end{equation}
where $H$ represents the cross entropy loss. $P_{emo}$ and $P_{id}$ are the ground truth categories for emotion and speaker id, while $Q_{emo}$ and $Q_{id}$ are the predicted categories for emotion and speaker id. 
Then, we freeze the parameters of the classifier and use its predictions as a guide to learn the style vector for the cascaded gesture synthesizer.

\subsection{HyperNetwork}
Since there exists a continuous spectrum of similarity among different emotions and speakers, the feature space inherently contains local neighborhood structure.
For example, emotions like happy, sad, angry, etc. exist on a spectrum from positive to negative valence. Speaking styles like tone of voice, pitch, emphasis, etc. also vary continuously rather than being completely distinct types.
To effectively capture the correlations between different representations, we employed a temporal convolutional network to model the style vector.
Specifically, two embeddings are employed to transform the speaker and emotion labels into vector representations. We employ a two-layer temporal convolutional network as a hypernetwork to generate a style vector that represents the style of gestures.
The style vector is then mapped into channel-wise mean and standard deviation parameters, which serve to modulate the latent gesture feature.

\subsection{The cascade of body generator and hand generator}
We cascade the body generator and hand generator by concatenating the features from the previous module and using them as inputs for subsequent decoders to extract latent features. This allows the features generated by the body generator to also be used in the hand generator. 

\subsection{About the user study}
We differentiated between expressive and non-expressive gestures when selecting speech-gesture combinations for the user study. The highly expressive combinations, like ``wow" and ``long", received more positive responses from participants compared to less expressive ones. This suggests that users recognized and responded to the differences in expressiveness. Additionally, it's important to note that our user study did not exclusively focus on expressive combinations. Instead, we aimed to achieve a balanced random selection of speech-gesture combinations based on the ground truth annotations of emotional categories derived from the generated results.
\section{Further Analysis}
\subsection{Ablation Study of Hyper-parameters}
Given the impact of hyperparameters on method performance, we conducted an ablation study on the hyperparameters used for computing the loss. As shown in Table~\ref{table:ablation_supp}, we find that assigning a larger weight to $\lambda_{\text{Mse}}$ and a smaller weight to $\lambda_{\text{Rhy}}$ benefits the learning of the model. The reason for this is that the optimization of rhythmic identification loss relies on facial features, so the model must focus on learning how to convert speech into facial blendshapes first. When the model becomes proficient in this aspect (e.g., when the value of $\lambda_{\text{Mse}}\cdot\ell_{\text{Mse}}$ becomes small), the model's learning of the rhythm will be accelerated.
\begin{table}[b]
\centering
\caption{Ablation study on hyper-parameters of our proposed method. $\downarrow$ denotes the lower the better, and $\uparrow$ denotes the higher the better.}
\label{table:ablation_supp}
\footnotesize
\begin{tabular}{ccccccc}
\toprule
\begin{tabular}[c]{@{}c@{}}($\lambda_{\text{Rec}},\lambda_{\text{Mse}},\lambda_{\text{Rhy}}$)\end{tabular} & FGD$\downarrow$ & SRGR$\uparrow$ & BeatAlign$\uparrow$ \\ 
\midrule
$(500,500,500)$  & 58.67 & 0.297 & 0.924   \\
$(1000,500,500)$  & 64.32 & 0.290 & 0.924   \\
$(1000,500,100)$  & 56.21 & 0.283 & 0.925  \\
$(500,1000,100)$  & 58.75 & 0.296 & 0.926 \\
$(1000,500,10)$  & 46.31 & 0.297 & 0.925   \\
$(500,1000,10)$  & 50.52 & 0.295 & 0.943 \\
$(1000,500,1)$  & 48.29 & 0.290 & 0.927 \\
$(500,1000,1)$  & 45.87 & 0.308 & 0.931 \\
\bottomrule
\end{tabular}%
\end{table}

\subsection{The influence of different modalities}
\begin{figure}[htp!]
\begin{center}
\includegraphics[width=1.0\linewidth]{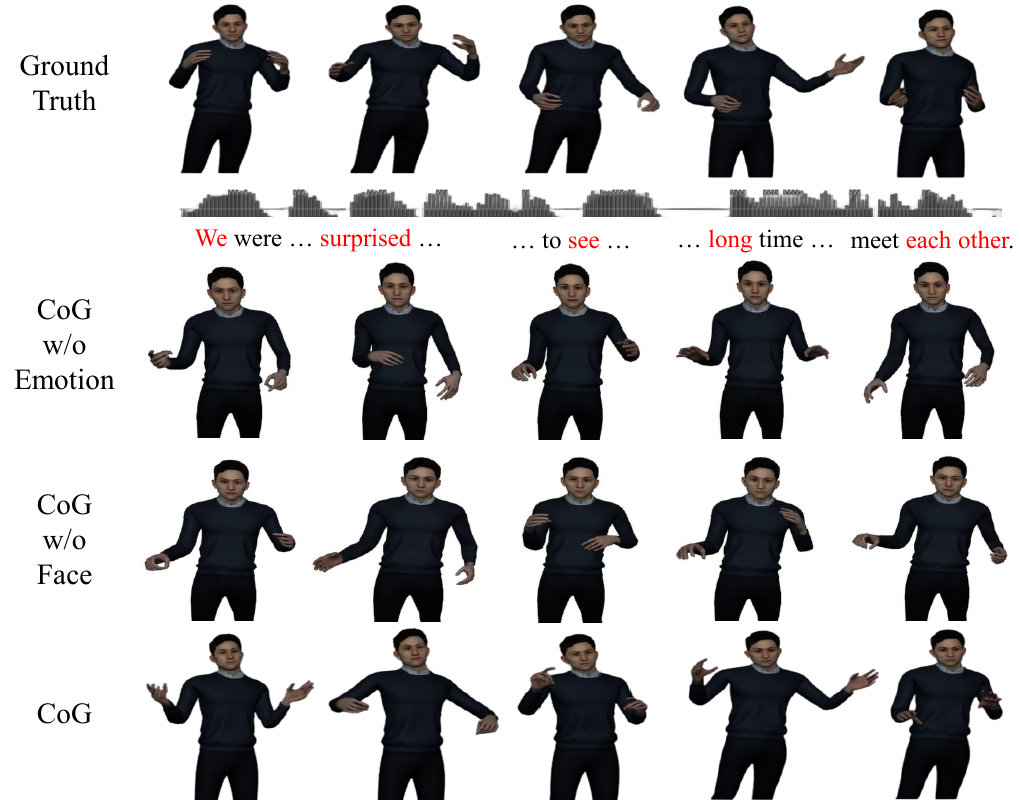}
\end{center}
   \caption{Visualization of the gestures generated by our method 
       without/with  different modalities.
   } 
\label{fig:sup1}
\end{figure}

As shown in figure~\ref{fig:sup1}, The incorporation of various modalities provides distinct benefits to the generation process. For instance, facial modality features can contribute to better capturing the rhythmic aspect of gestures. On the other hand, style features from the emotional modality can enable more personalized representations, avoiding a generic or averaged outcome.

\subsection{More Subjective Results}
\begin{figure*}
\begin{center}
\includegraphics[width=1.0\linewidth]{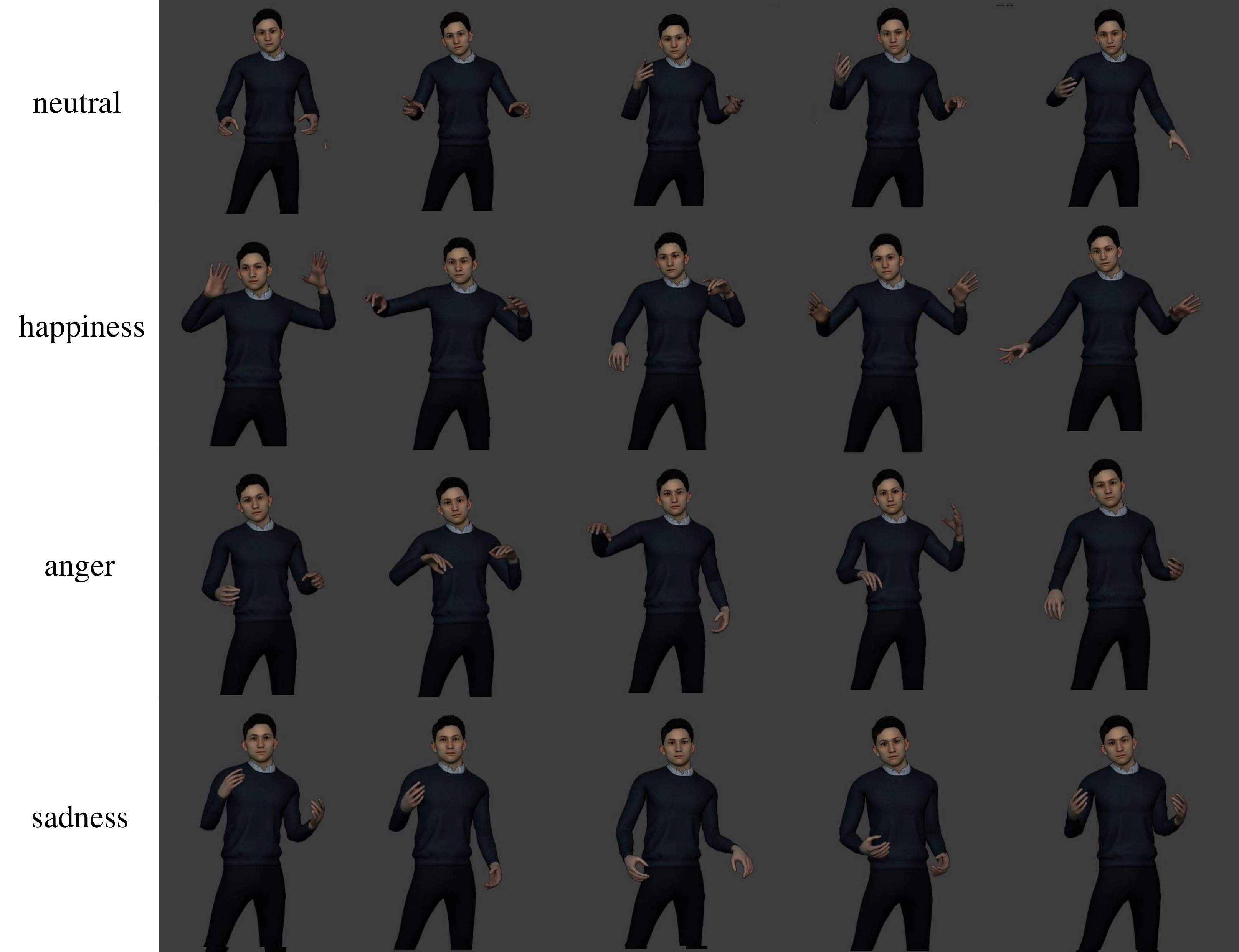}
\end{center}
   \caption{Visualization of 3D gestures generated in different emotional states. The results of different emotions are presented in separate rows, with each row representing the results of a specific emotion at different time frames. 
} 
\label{fig:emo1}
\end{figure*}

\begin{figure*}
\begin{center}
\includegraphics[width=1.0\linewidth]{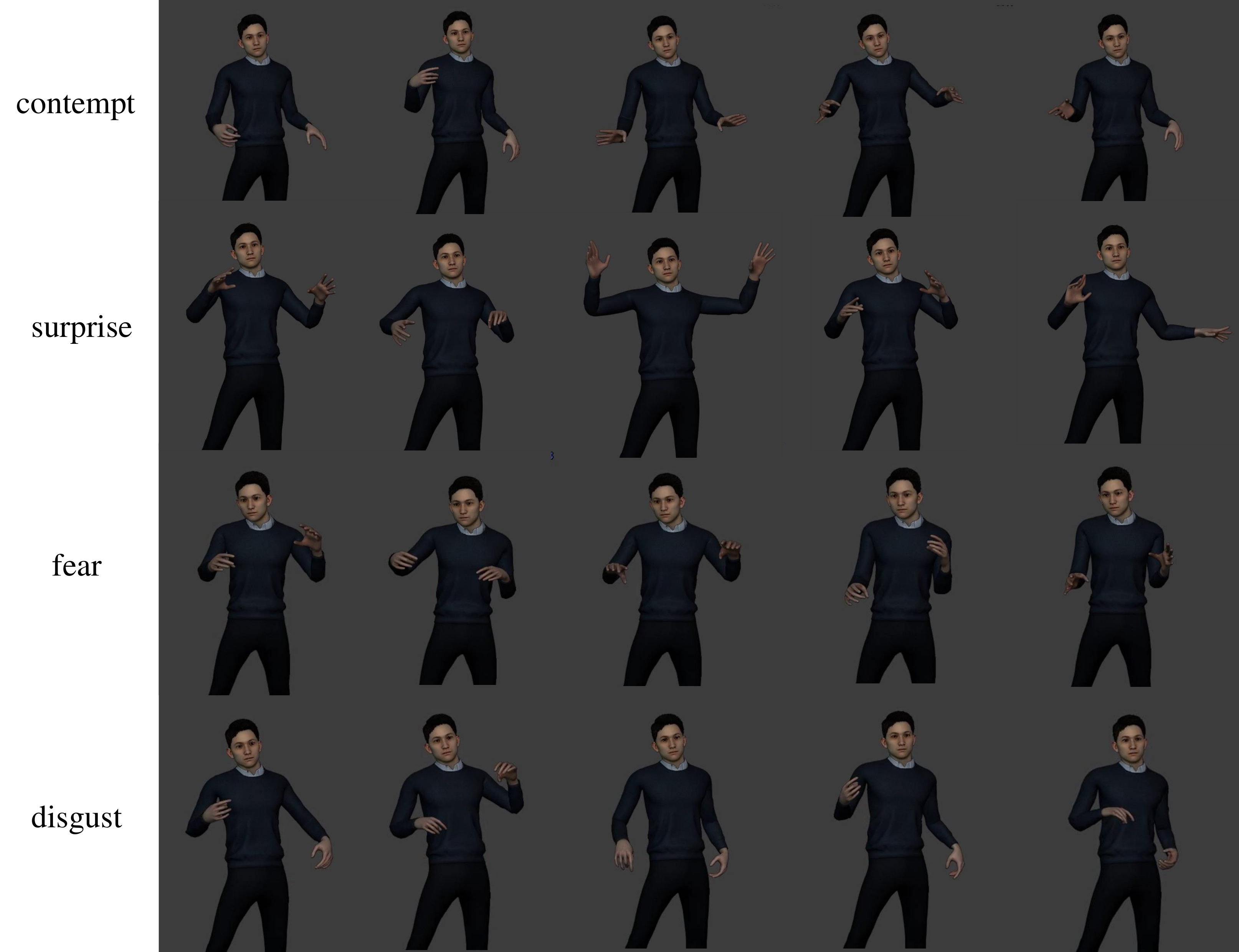}
\end{center}
   \caption{Visualization of 3D gestures generated in different emotional states. The results of different emotions are presented in separate rows, with each row representing the results of a specific emotion at different time frames.
} 
\label{fig:emo2}
\end{figure*}

To better demonstrate our method's ability to generate gestures across different styles, we further showcase additional results conditioned on speech with diverse emotions and speakers.
As shown in Figure~\ref{fig:emo1} and Figure~\ref{fig:emo2}, we observe clear variations in the gestures produced during different emotions. For example, when experiencing ``happiness'' or ``surprise'', gestures tend to expand outward and change at a faster pace. On the other hand, when feeling ``fear'' or ``sadness'', gestures tend to become more inward and have fewer variations. Additionally, we observe that similar emotions also lead to more similar gestures, such as ``contempt'' and ``disgust'', ``surprise'' and ``happiness'', etc.

\begin{figure*}
\begin{center}
\includegraphics[width=1.0\linewidth]{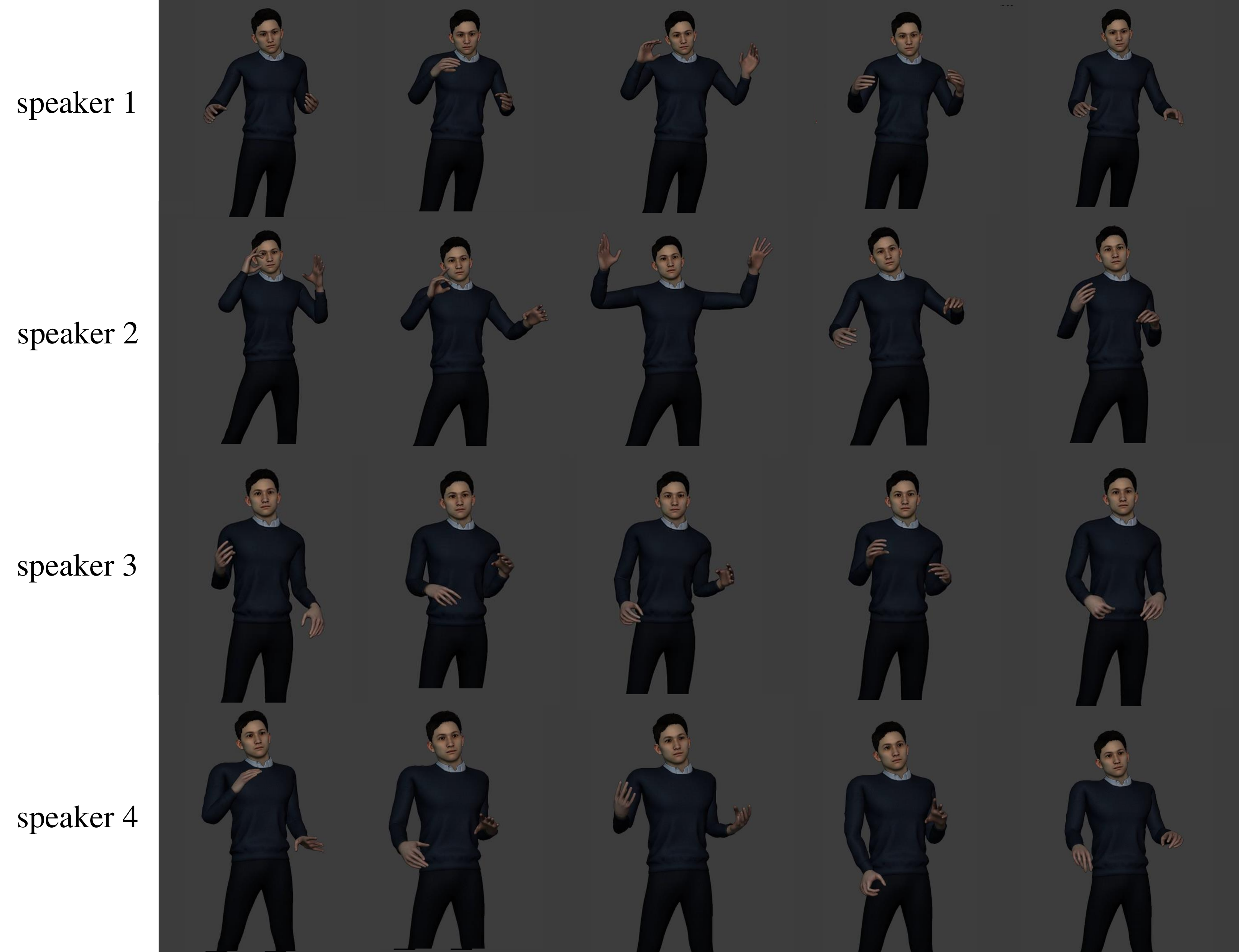}
\end{center}
   \caption{Visualization of 3D gestures generated by different speakers. The results of different speakers are displayed in separate rows, with each row representing the results of a specific speaker at different time  frames.
} 
\label{fig:spk1}
\end{figure*}

\begin{figure*}
\begin{center}
\includegraphics[width=1.0\linewidth]{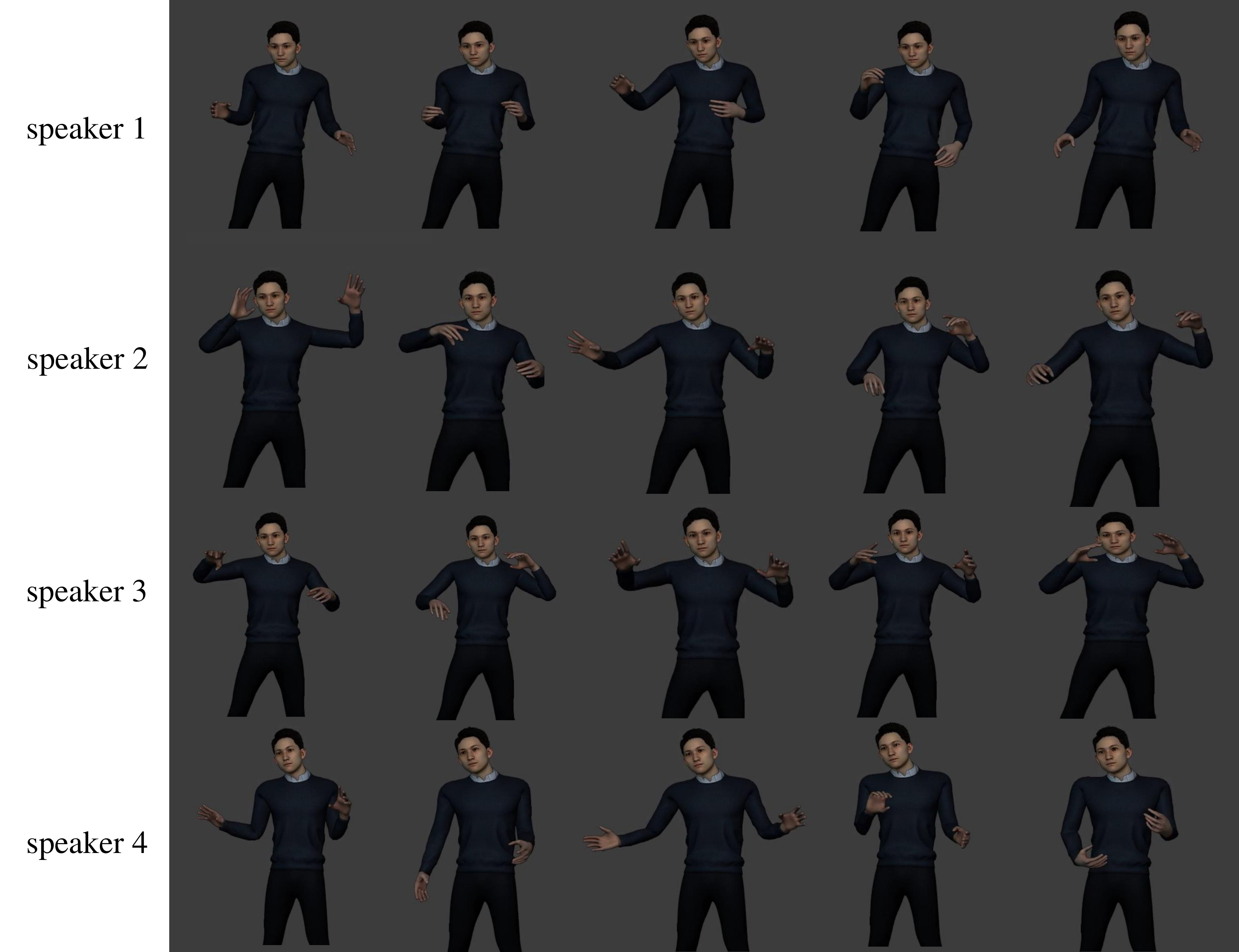}
\end{center}
   \caption{Visualization of 3D gestures generated by different speakers. The results of different speakers are displayed in separate rows, with each row representing the results of a specific speaker at different time  frames.
} 
\label{fig:spk2}
\end{figure*}

From Figure~\ref{fig:spk1} and Figure~\ref{fig:spk2}, it can be observed that the generated gestures vary across different speakers. To control for other factors, we kept the speech content the same within a single image. In Figure~\ref{fig:spk1}, it can be seen that gestures from speaker 2 tend to expand outward with a larger magnitude of variations, while gestures from speaker 3 and speaker 4 tend to contract inward with relatively smaller variations. This observation is partially related to the gender of the speakers (e.g., speaker 3 and speaker 4 are female, while speaker 1 and speaker 2 are male). However, for the speech content corresponding to Figure~\ref{fig:spk2}, each speaker has a unique gestural style, and there are no significant differences in gesture styles based on gender. 

\end{document}